\documentclass{article}

\usepackage{arxiv}

\usepackage{amsmath}
\usepackage{amsthm}
\usepackage[utf8]{inputenc} 
\usepackage[T1]{fontenc}    
\usepackage{hyperref}       
\usepackage{url}            
\usepackage{booktabs}       
\usepackage{amsfonts}       
\usepackage{nicefrac}       
\usepackage{microtype}      
\usepackage{lipsum}		    
\usepackage{graphicx}
\usepackage{doi}
\usepackage{tikz}
\usepackage{subcaption}
\usepackage{float}
\usepackage{wrapfig}
\usepackage{algpseudocode}
\usepackage{algorithm}
\usepackage{algorithmicx}

\usepackage{pgfplots}
\usepgfplotslibrary{fillbetween}
\usepackage{filecontents}

\newtheorem{theorem}{Theorem}
\newtheorem{lemma}[theorem]{Lemma}
\newtheorem{remark}[theorem]{Remark}

\definecolor{c_red}{RGB} {246,182,186}
\definecolor{c_blue}{RGB} {178,218,233}
\definecolor{c_green}{RGB} {160,202,121}

\title{Efficient Encoding of Graphics Primitives with Simplex-based Structures}

\newcommand*\samethanks[1][\value{footnote}]{\footnotemark[#1]}

\author{{Yibo Wen \thanks{$ $ indicates equal contribution.}}\\
	Northwestern University\\
    \href{mailto:yibowen2024@u.northwestern.com}{\texttt{yibowen2024@u.northwestern.com}} \\
	\And
	{Yunfan Yang \samethanks} \\
	Northwestern University\\
 \href{mailto:frankyang2024@u.northwestern.edu}{\texttt{frankyang2024@u.northwestern.edu}} \\
}

\date{}

\renewcommand{\shorttitle}{Efficient Encoding of Graphics Primitives with Simplex-based Structures}

\hypersetup{
pdftitle={Efficient Encoding of Graphics Primitives with Simplex-based Structures},
pdfauthor={Yibo Wen, Yunfan Yang},
pdfkeywords={Simplex Structures, Graphics Primitives, Encodings, Date Compression, Parallel Computating},
}

\begin{document}

\maketitle

\begin{abstract}
    Grid-based structures are commonly used to encode explicit features for graphics primitives such as images, signed distance functions (SDF), and neural radiance fields (NeRF) due to their simple implementation. However, in $n$-dimensional space, calculating the value of a sampled point requires interpolating the values of its $2^n$ neighboring vertices. The exponential scaling with dimension leads to significant computational overheads. To address this issue, we propose a simplex-based approach for encoding graphics primitives. The number of vertices in a simplex-based structure increases linearly with dimension, making it a more efficient and generalizable alternative to grid-based representations. Using the non-axis-aligned simplicial structure property, we derive and prove a coordinate transformation, simplicial subdivision, and barycentric interpolation scheme for efficient sampling, which resembles transformation procedures in the simplex noise algorithm. Finally, we use hash tables to store multiresolution features of all interest points in the simplicial grid, which are passed into a tiny fully connected neural network to parameterize graphics primitives. We implemented a detailed simplex-based structure encoding algorithm in C++ and CUDA using the methods outlined in our approach. In the 2D image fitting task, the proposed method is capable of fitting a giga-pixel image with 9.4\% less time compared to the baseline method proposed by instant-ngp, while maintaining the same quality and compression rate. In the volumetric rendering setup, we observe a maximum 41.2\% speedup when the samples are dense enough.
\end{abstract}


\section{Introduction}
Graphics primitives are building blocks used to create complex visual scenes in computer graphics. These primitives can be thought of as functions that map positional or directional information from $\mathbb{R}^m$ to attributes in $\mathbb{R}^n$. The quality and performance characteristics of the mathematical representation are crucial for visual fidelity: we desire representations that remain fast and compact while capturing high-frequency, local detail. Recently, we have seen a trend in representing graphics primitives with neural networks as demonstrated by advancements in occupancy networks~\cite{Occupancy_Networks, chen2018implicit_decoder}, signed distance functions~\cite{Park_2019_CVPR, Vicini2022sdf} and radiance fields\cite{mildenhall2020nerf}. While neural networks such as multi-layer perceptrons (MLPs) have shown great potential in modeling graphics primitives, they struggle to capture high-frequency details due to their inherent smoothness.

To overcome this limitation, neural networks are often paired with various encoding techniques, which map input into higher dimensionality to allow a finer representation of high-frequency details. Implicit encodings, such as frequency encoding used by authors in NeRF~\cite{mildenhall2020nerf}, encodes scalar positions as a multi-resolution sequence of sine and cosine functions. These type of encodings normally don't carry any trainable parameters and therefore require a larger neural network to achieve the same level of fidelity. Conversely, explicit encodings rely on additional grid-based structures with trainable parameters. Recent examples of grid representation includes dense grid~\cite{SunSC22, Lombardi:2019}, sparse grid~\cite{hedman2021snerg, yu_and_fridovichkeil2021plenoxels}, octree~\cite{yu2021plenoctrees}, tensor decomposition~\cite{Chen2022ECCV}, planar factorization~\cite{cao2023hexplane, Chan2021, fridovich2023k}, and multi-resolution grid~\cite{mueller2022instant}. These configurations aim to exchange a reduced memory footprint for a lower computational cost with the use of a smaller neural network.

Nonetheless, incorporating grid-based structures inevitably introduces excessive computation, as evaluating an input in $n$ dimensions necessitates evaluating $2^n$ neighboring vertices. Drawing inspiration from simplex noise, we leverage simplex structures, defined as the polygon with the minimum number of vertices tiling an n-dimensional space. The use of simplex-based structures proves more advantageous than grid-based structures in encoding for two main reasons:

\textbf{Fewer variables} In dense-grid encoding, the number of variables required increases exponentially with the dimensionality of the graphics primitives, making it impractical for high-dimensional problems. In contrast, a simplex-based structure uses only the n+1 vertices in n-dimension regardless of the dimensionality. This would lead to a significant improvement in computation speed.

\textbf{Fewer artifacts} In simplex-based structure, the vertices of the simplex are typically well-separated and represent different combinations of variables. This reduces the correlation between variables, which in turn reduces the likelihood of artifacts arising due to the interactions between variables. Simplex-based encoding can more easily handle nonlinearities due to its ability to adapt to the shape of the solution space. In contrast, the dense-grid shape is fixed, leading to artifacts on discontinuities or sharp edges.

In later sections, we will further review the properties of grid and simplex structures on noises (Section 3) and present our proposed method with simplex-based encoding (Section 4) with the implementation of a simplex-based structure backed by multi-resolution hash encoding, a state-of-the-art method on graphics primitives (Section 5). We then verify our performance and feasibility with multi-dimensional experiments on various tasks (Section 6). We finally conclude with future works and discussions thereof (Section 7) with mathematical derivations of the proposed algorithm feasibility (Section 8).

\section{Background}

\subsection{Perlin noise} 
Perlin noise~\cite{perlin_noise}, also known as classical noise, is a procedural generation algorithm first introduced by Ken Perlin in the 1980s. Due to its natural appearance and simple implementation, it has been widely adopted in computer graphics for generating visual content including texture, terrain, smoke, etc.
\label{sec:headings}

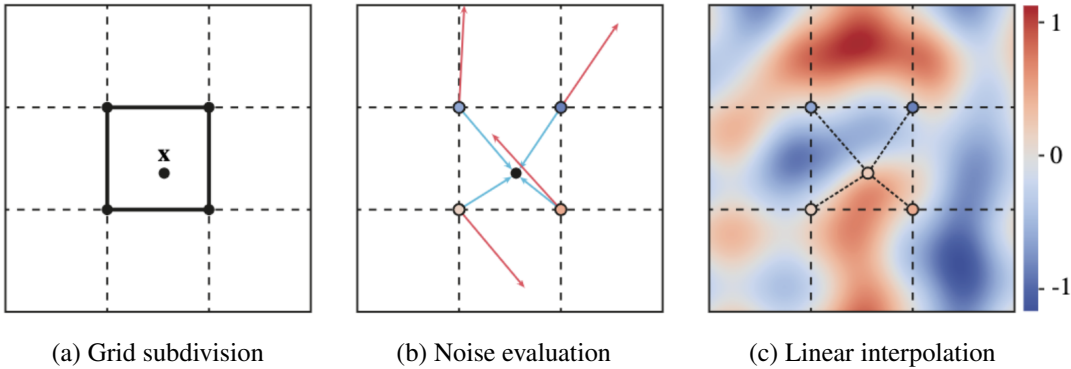
\begin{figure}[h]
\centering
\begin{tikzpicture}
    \draw (0, 0) node[inner sep=0] {\includegraphics[width=.9\linewidth]{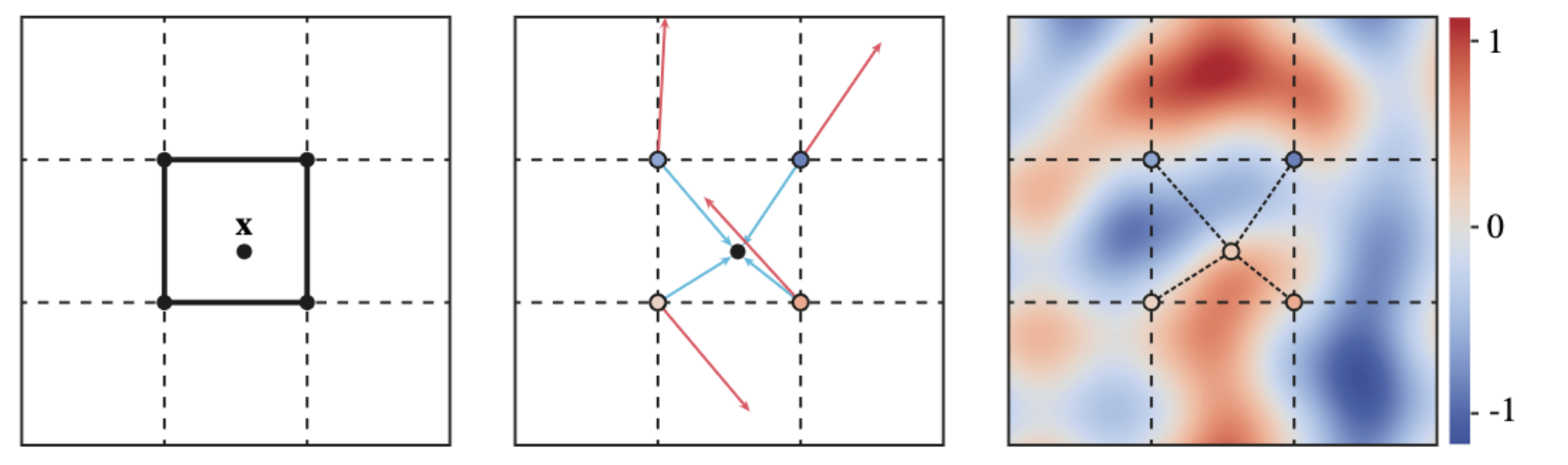}};
    
    \draw (-5.2, -2.6) node {(a) Grid subdivision};
    \draw (-0.6, -2.6) node {(b) Noise evaluation};
    \draw (4.3, -2.6) node {(c) Linear interpolation};
\end{tikzpicture}
\caption{Illustration of Perlin noise in 2D.}
\label{fig:perlin}
\end{figure}

In $n$ dimensional space, Perlin noise can be viewed as a pseudo-random mapping $\mathbb{R}^n \to \mathbb{R}$ and can be calculated with the following three steps as shown in Figure \ref{fig:perlin}: (a) Grid subdivision. Given an input coordinate $\textbf{x} \in \mathbb{R}^n$, we determine a grid cell that contains $\textbf{x}$. The cell is a $n$ dimensional hypercube with $2^n$ vertices in $\mathbb{Z}^n$ spanned by $\lfloor \textbf{x} \rfloor$ and $\lceil \textbf{x} \rceil$. (b) Noise evaluation. For each vertex $\textbf{x}_i \in \mathbb{Z}^n$ of the hypercube, we generate a pseudo-random gradient vector of unit length $\textbf{g}_i \in \mathbb{R}^n, ||\textbf{g}_i||=1$ and a displacement vector $\textbf{d}_i = \textbf{x} - \textbf{x}_i$. We then use the dot product $p_i = \textbf{g}_i \cdot \textbf{d}_i$ of the two vectors as the noise vector. (c) Linear interpolation. Finally, we use $n$-linear interpolation to obtain the final noise scalar at $\textbf{x}$. When higher-order smoothness is desired, instead of using $n$-quadratic  or $n$-cubic interpolation, we can apply a more efficient smoothing function to the $2^n$ dot products on vertices. Particularly, Perlin noise uses the smoother-step function, \begin{equation} \label{e1}
S_2(x) = 6x^5 - 15x^4 + 10x^3, 0\leq x \leq 1, 
\end{equation} which is $C^2$ smooth and has vanishing derivatives at the endpoints. Considering the procedures mentioned above, Perlin noise requires evaluation at $2^n$ vertices of the containing grid cell and calculation of $2^{n-1}$ weighted sums during interpolation. Therefore, the algorithm scales with $\mathcal{O}(2^n)$, which grows exponentially with dimension.

\subsection{Simplex noise}
Perlin noise, while very useful, suffers from exponential scaling across dimensions and directional artifacts in image reconstruction \cite{simplex_noise}. Those shortcomings inspire us to investigate a new reconstruction primitive: simplex noise. Rather than placing each input point into a cubic grid based on the integer parts of coordinate values, the input point is placed onto a simplicial grid, which is derived by dividing n-dimensional space into a regular grid of shapes with a minimum number of vertices (triangles in 2D, tetrahedrons in 3D, and so on). It's important to note that the number of simplex vertices in each dimension is $n+1$, where $n$ is the number of sizes.

Compared to Perlin noise, simplex noise can be generated with lower computational overhead, especially in higher dimensions. Simplex noise scales to higher dimensions with much less computational cost: the complexity is $\mathcal{O}(n^2)$ (depending on the sorting algorithm)  for $n$ dimensions. Simplex noise inherently produces fewer noticeable directional artifacts and is more isotropic (meaning it looks the same from all directions) than the Perlin noise. Both of these advantages over Perlin noise are crucial for a robust and computationally efficient image reconstruction algorithm.

\begin{figure}[h]
\centering
\begin{tikzpicture}
    \draw (0, 0) node[inner sep=0] {\includegraphics[width=0.7\linewidth]{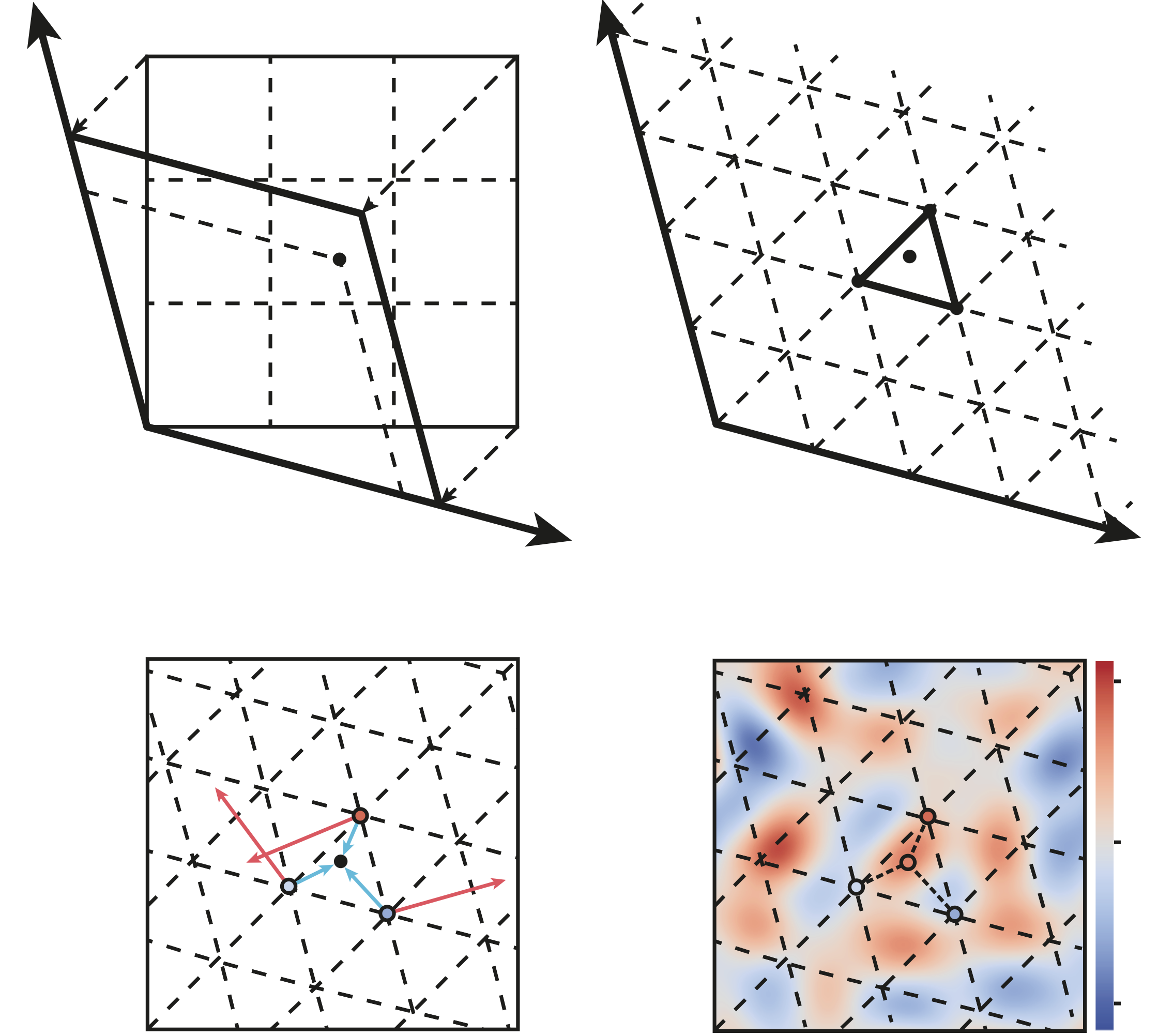}};
    \draw (-2.42, 2.8) node {$\textbf{x}$};
    \draw (5.55, -1.61) node {1};
    \draw (5.55, -3.21) node {0};
    \draw (5.62, -4.80) node {-1};
    
    \draw (-2.5, -0.9) node {(a) Coordinate skewing};
    \draw (3.3, -0.9) node {(b) Simplicial subdivision};
    \draw (-2.5, -5.5) node {(c) Noise evaluation};
    \draw (3.3, -5.5) node {(d) Kernel summation};
\end{tikzpicture}
\caption{Illustration of Simplex noise in 2D.}
\label{fig:simplex}
\end{figure}
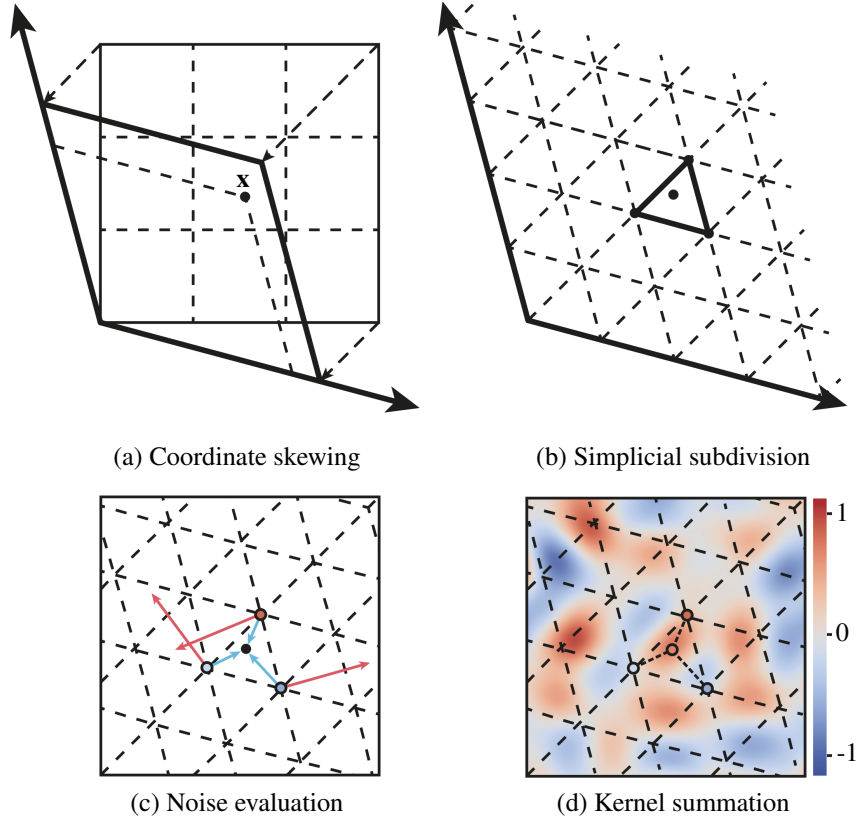

To generate a simplex noise in $n$ dimensional space, the following 4 steps must be made as shown in Figure \ref{fig:simplex}.

\textbf{(a) Coordinate skewing:} The coordinate axis in the $n$ dimension is skewed such that the coordinate vector aligns with the simplex shape. In a 2D example, the x-y Cartesian coordinate is translated to a new u-v plane. The coordinate translation formula is given below,
\begin{equation} \label{e2}
\textbf{x}^\prime = \textbf{x} + \textbf{1}_n^T \cdot F_n \sum_i{x_i},\quad F_n = \frac{\sqrt{n+1}-1}{n}, 
\end{equation}

This has the effect of rearranging a hyper-cubic coordinate that has been squashed along its main diagonal such that the distance between the points (0, 0, ..., 0) and (1, 1, ..., 1) becomes equal to the distance between the points (0, 0, ..., 0) and (1, 0, ..., 0). 

\textbf{(b) Simplicial subdivision:} Once the input coordinate is determined in the translated coordinate system, the surrounding lattice point of an input in the simplex grid is calculated via the following steps. First, take a floor and ceiling of the coordinates in the input. For input with coordinate ($x,y,z,...$) in the simplex coordinate, it lies in a simplex with at least coordinate spanned by ($\lfloor \textbf{x} \rfloor, \lfloor \textbf{y} \rfloor, \lfloor \textbf{z} \rfloor...)$ and ($\lceil \textbf{x} \rceil, \lceil \textbf{y} \rceil, \lceil \textbf{z} \rceil...)$. Then, the coordinate ($x_i, y_i, z_i, ...$) are sorted in decreasing order. Start with ($\lfloor \textbf{x} \rfloor, \lfloor \textbf{y} \rfloor, \lfloor \textbf{z} \rfloor...)$, successively add 1 to the largest point in the coordinate until all $n+1$ simplex points are found. 

\textbf{(c) Noise evaluation:} At each vertex of the grid, a random gradient vector is assigned. To generate a noise value at a given point in space, the algorithm first determines which simplex shape contains the point, and then interpolates the gradient vectors at the vertices of that simplex to obtain a weighted sum. The resulting value is then scaled and smoothed to produce the final noise value.

\textbf{(d) Kernel summation:} The input in the simplex coordinate is skewed back into the Cartesian coordinate system using the formula below,

\begin{equation} \label{e3}
\textbf{x} = \textbf{x}^\prime - \textbf{1}_n^T \cdot G_n \sum_i{x_i^\prime},
    \quad G_n = \frac{1-1/\sqrt{n+1}}{n}.
\end{equation}

Note that the translated unscrewed coordinate is precisely the coordinate of the original input in the orthogonal axes. 

\section{Proposed Method}

To optimize the performance of sampling and interpolation, it may be beneficial to replace $n$-cubes with $n$-simplices, as simplices are defined as polygons with the least number of vertices in each respective dimension. For example, a simplex is a triangle in two-dimensional space and a tetrahedron in three-dimensional space. However, regular simplices that have equal edges cannot tile space beyond two dimensions. Furthermore, indexing the vertices of equilateral triangles in two dimensions is no easy task without careful manipulation. Fortunately, the simplex noise algorithm has provided a solid foundation for such operations. By making some adaptations, we can apply this methodology to a wider range of tasks, including parameterizing graphics primitives. In doing so, we can make full use of simplex-based structures to reduce computational costs and optimize memory usage during sampling and interpolation. For showing the correctness of our proposed method in arbitrary dimensions, we derive and prove the following theorems and lemmas (please refer to \textbf{Appendix A} for detailed proofs):

\begin{figure}[h]
\centering
\begin{tikzpicture}
    \draw (0, 0) node[inner sep=0] {\includegraphics[width=0.8\linewidth]{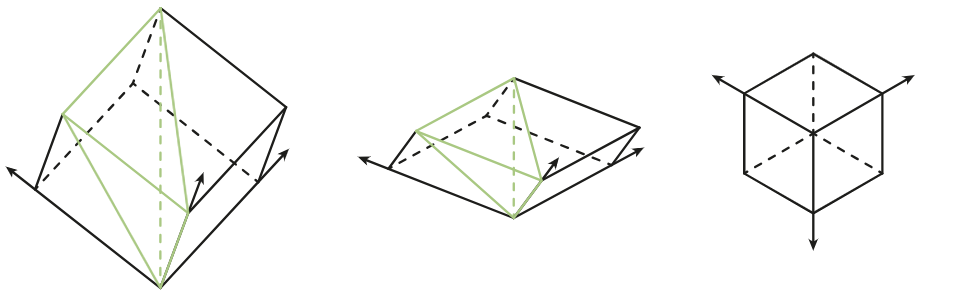}};
    \draw (-4.5, -2.8) node {(a) 3-orthoscheme};
    \draw (0.5, -2.8) node {(b) Coordinate skewing};
    \draw (4.8, -2.8) node {(c) Simplicial subdivision};
\end{tikzpicture}
\caption{{Illustration of the proposed method with coordinate skewing and simplicial subdivision in three dimensions.}}
\label{fig:skew}
\end{figure}
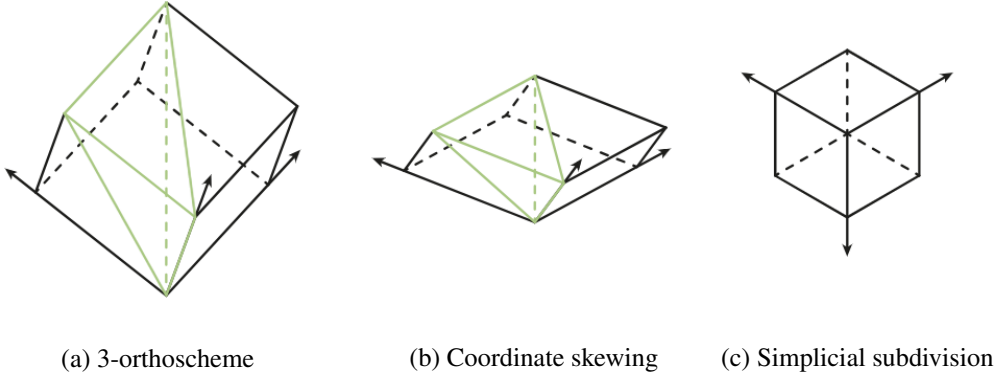

\textbf {Theorem 1. }$S=\{ \textbf{x}\in\mathbb{R}^n : 0\leq x_1\leq \cdots \leq x_n\leq 1\}$ is a $n$-simplex.

\textbf {Remark 2. }Let $\pi$ denote a permutation of $\{1,\cdots,n\}$, then $S_{\pi}=\{ \textbf{x}\in\mathbb{R}^n : 0\leq \textbf{x}_{\pi(1)}\leq \cdots \leq \textbf{x}_{\pi(n)}\leq 1\}$ is a $n$-simplex. The $n$-simplex has $n+1$ vertices $\textbf{v}_1, \cdots, \textbf{v}_{n+1}$, where all entries at $\{\pi_1,\cdots,\pi_{i-1}\}$ of $\textbf{v}_i$ are 1 and the rest are 0. Additionally, $\textbf{v}_1$ is $\textbf{0}^T$ and $\textbf{v}_{n+1}$ is $\textbf{1}^T$.

\textbf {Theorem 3. }All $S_{\pi}$ are congruent.

\textbf{Lemma 4. }After coordinate transformation, all possible $S_{\pi}^{\prime}$ are still congruent.

\textbf {Theorem 5. }A $n$-hypercube can be triangulated into $n!$ disjoint congruent simplices.

\subsection{Coordinate skewing}
Similar to the simplex noise algorithm, we first apply coordinate transformation according to Equation \ref{e2}. This skewing operation can also be rewritten as

\begin{equation} \label{e5} \textbf{x}^\prime = \begin{bmatrix} 
    1+F_n & \cdots  & 1\\
    \vdots & \ddots & \vdots\\
    1 & \cdots  & 1+F_n 
  \end{bmatrix} \textbf{x}, \quad F_n = \frac{\sqrt{n+1}-1}{n},
\end{equation}

where the sampled point $\textbf{x}\in \mathbb{R}^n$ is multiplied by a constant matrix. As such, coordinate skewing is an affine transformation \footnote{A map f: X → Z is an affine map if there exists a linear map $m_f$ : V → W such that $m_f$(x - y) = f (x) - f (y) for all x, y in X. Such transformation preserves the lines and parallelism.} that preserves the parallelism of planes and transforms the unit $n$-hypercube into an $n$-parallelpiped.

This transformation is crucial as it reduces distortion in simplex cells, resulting in the more balanced spatial division. Without this step, each cell would be a $n$-orthoscheme, which is a generalization of a right triangle in higher dimensions. The transformation leads to equilateral triangles in two-dimensional space and tetrahedrons with congruent isosceles faces in three-dimensional spaces. More specifically, this tetrahedron has 4 edges of the same length and another 2 edges of the same length which can be derived from the proof for Lemma~\ref{l1}. We can also calculate the ratio of the two different types of edges which is $\frac{\sqrt{3}}{2}$. Thus by using a coordinate transformation, we aim to avoid axis-aligned artifacts and increase quality in representing graphics primitives.

\subsection{Simplicial subdivision}

After transforming the input coordinate into simplex-based coordinates, we will need to locate the coordinates of neighboring vertices in the n-dimension hypercube. According to Theorem~\ref{t4}, in $n$-dimensional space, the hypercube can be split into $n!$ number of disjoint and congruent simplices. While congruency is not preserved under general affine transformation, we prove in Lemma~\ref{l1} that under our coordinate skewing, the $n!$ simplices are still congruent. Assume the given point is sampled within the hypercube, then to locate the cell containing it, we need to determine which simplex cell among the $n!$ simplices it resides. For this step, we use a similar subdivision scheme from the simplex noise algorithm. Through direct sorting, we can find the corresponding vertices of the cell as shown in the proof for Remark~\ref{t2}. For a given point $\textbf{x}$ inside the unit hypercube, we let $x_1, \cdots, x_n$ denote the sorted entries of $\textbf{x}$ in descending order and record their respective original index. Coding-wise, we perform sorting for the input while preserving their index \footnote{Bubble sort is used since $n$ is small, under which bubble sorting outperforms other methods.}. Then, we can obtain the $n+1$ vertices starting from $\textbf{0}^T$ as the base vertex $\textbf{v}_1$ \footnote{After scale adjustment, all input coordinates are less than 1. $\textbf{0}^T$ is guaranteed to be the lower corner of the simplex structure.}. By adding 1 to the base vertex at the index of the next largest entry of $\textbf{x}$, we obtain the next vertex and use that as the new base. The process is repeated until we finally reach vertex $\textbf{v}_{n+1}$, which should always be $\textbf{1}^T$.

\subsection{Barycentric interpolation}

\begin{figure}[h]
\centering
\begin{tikzpicture}
    \draw (0, 0) node[inner sep=0]{\includegraphics[width=0.6\linewidth]{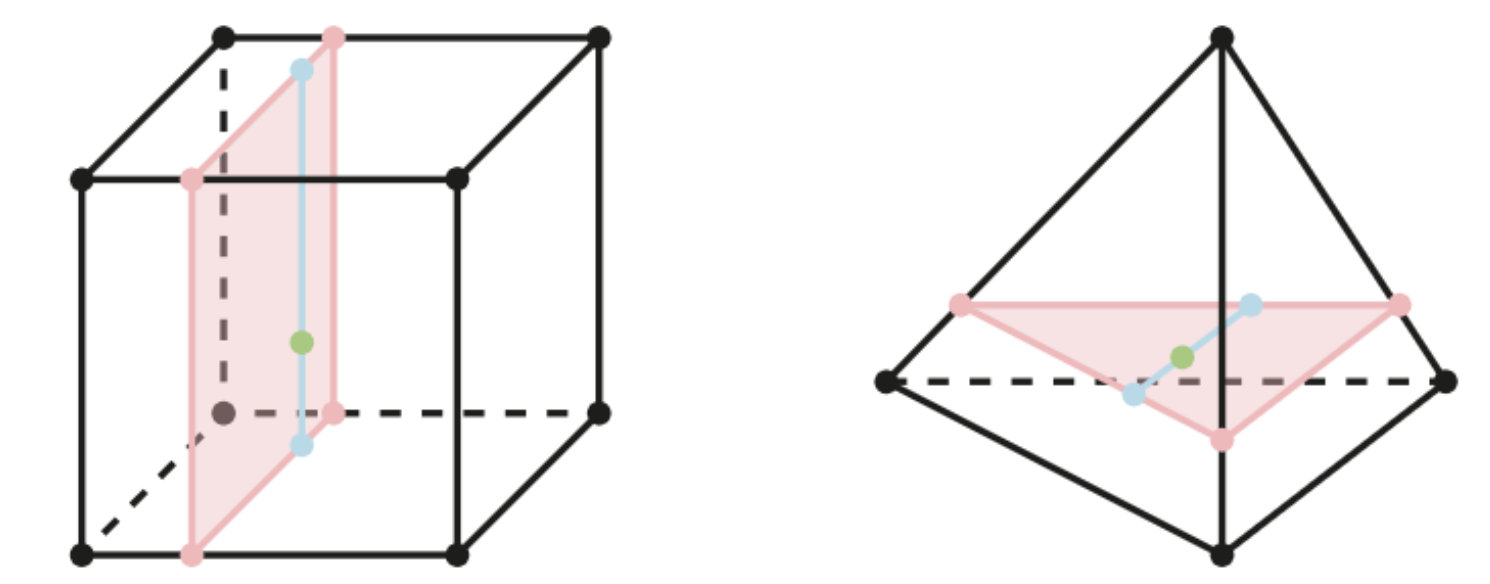}};
    \draw (-2.8, -3.0) node {(a) Trilinear interpolation};
    \draw (3.2, -3.0) node {(b) Barycentric interpolation};
\end{tikzpicture}
\caption{Illustration of linear interpolation for cube and tetrahedron.}
\label{fig:interp}
\end{figure}

In order to learn the mapping for graphics primitives, instead of generating random gradient vectors as in noise algorithms, we retrieve learnable vectors at each vertex and interpolate them to obtain the final feature for the queried point. To maintain a similar interpolation scheme as the original trilinear interpolation, we didn't select kernel summation from the simplex noise algorithm. Instead, we derive our barycentric interpolation as an efficient alternative which is illustrated in the Figure~\ref{fig:interp}.

In geometry, a barycentric coordinate system is a coordinate system in which the location of a point is specified by reference to a simplex, which makes it the perfect choice for our simplex-based structure. The barycentric coordinate can be found by expressing the point inside a simplex using a convex combination of the neighboring $n+1$ vertices and the coefficients of such combination are the local barycentric coordinate. Since the coordinate skewing is affine, the barycentric coordinate is preserved and the weights are given in the proof of Theorem~\ref{t1} as follows, 
\begin{equation}
    w_1 = 1-x_1, w_2 = x_1-x_2, \cdots, w_{n} = x_{n-1} - x_n, w_{n+1} = x_n,
\end{equation}

where the entries of $\textbf{x}$ are already sorted in descending order from the previous step. Compared to the computationally expensive $n$-linear interpolation, the use of a simpler formula for weights in our proposed algorithm also enables a more efficient implementation.

\subsection{Example Algorithm}

Here we provide an example pseudo-code of our proposed method. The function takes in a point $\textbf{x}$ inside the unit hypercube, retrieves values at its neighboring simplex vertices, and returns their interpolated values. With slight modifications, this example can be adapted using CUDA for more advanced scenarios including processing inputs in parallel, handling feature vectors, etc. For a detailed demonstration of Python and sample outputs for each phase of our proposed method, please refer to \textbf{Appendix B}. 

\section{Implementation}

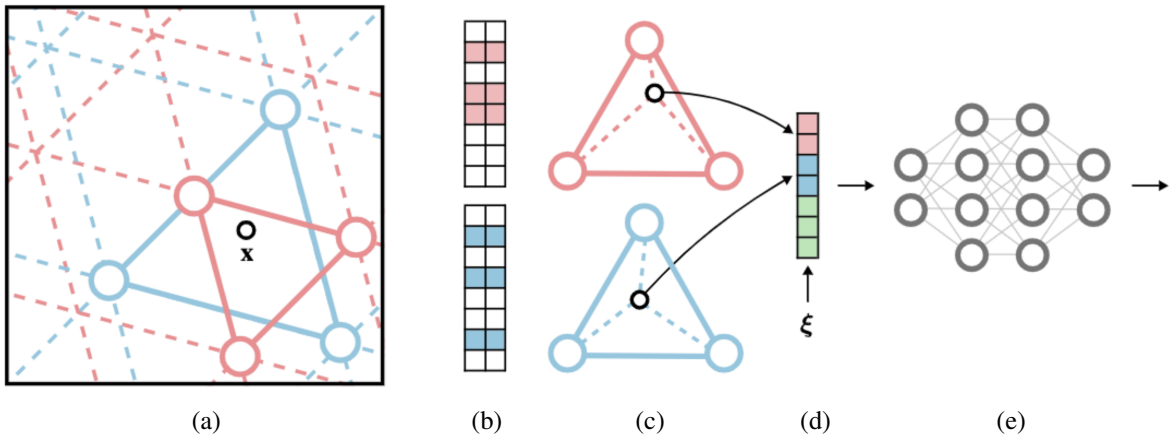
\begin{figure}[h]
\centering
\begin{tikzpicture}
    \draw (0, 0) node[inner sep=0] {\includegraphics[width=0.95\linewidth]{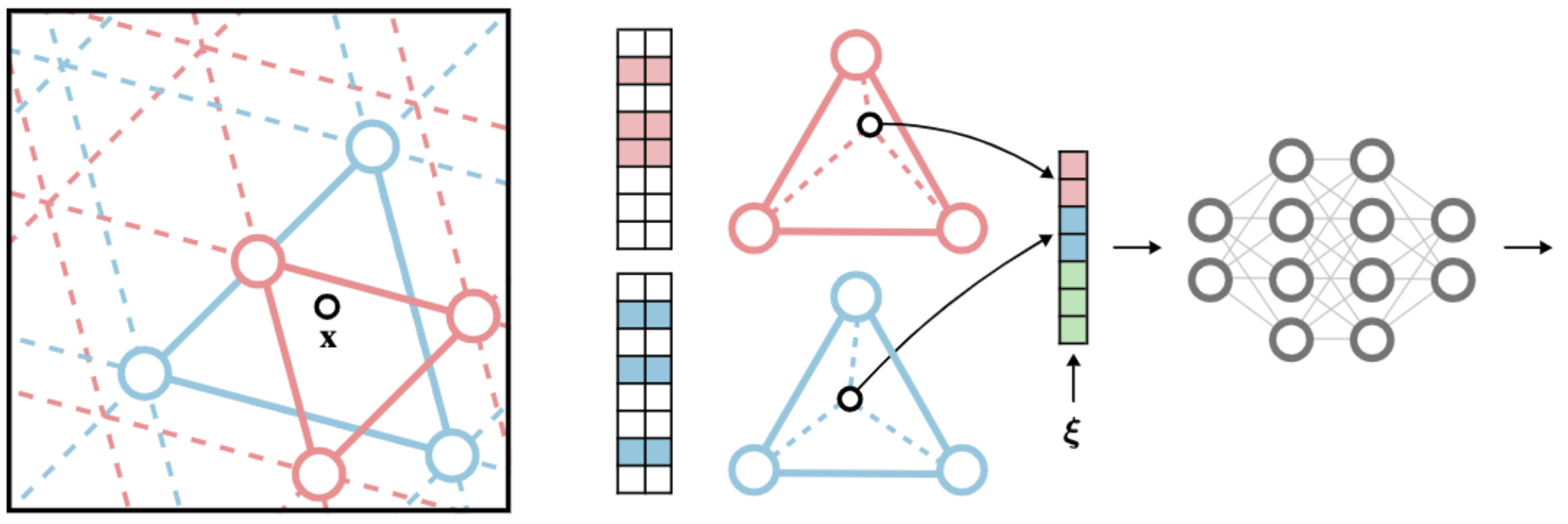}};
    
    \draw (-5.1, -3.0) node {(a)};
    \draw (-1.37, -3.0) node {(b)};
    \draw (0.8, -3.0) node {(c)};
    \draw (3.0, -3.0) node {(d)};
    \draw (5.6, -3.0) node {(e)};
\end{tikzpicture}
\caption{Illustration of the proposed method in 2D in combination with multi-resolutional hash encoding.}
\label{fig:instant-NGP}
\end{figure}

To validate our proposed structure for representing graphics primitives, we adopt the structural backbone from Instant-NGP~\cite{mueller2022instant} and replace its explicit grid-based structure with a simplex-based structure. The demonstration of our implementation in 2D is given in Figure \ref{fig:instant-NGP}: 

\textbf{(a) Hashing of voxel vertices} Find the surrounding neighbors at L different resolution levels on a 2D plane and assign indices to their corners by hashing their integer coordinates. 
\textbf{(b) Lookup} Look up the corresponding n-dimensional feature vectors from the hash tables $H_L$ for all resulting corner indices.
\textbf{(c) Barycentric Interpolation} Perform barycentric interpolation on neighboring coordinates according to the relative position of x within the respective l-th voxel. \textbf{(d) Concatenation} Concatenate the result of each level as well as any auxiliary inputs, producing the encoded MLP input. 
\textbf{(e) Neural Network } Backpropagate loss function through the MLP (e) the concatenation (d), the linear interpolation (c), and then accumulated in the looked-up feature vectors. 

\subsection{Scale Adjustment}

In the multiresolution setup, sampling still takes place inside the unit $n$-cube. The difference is that the $n$-cube is divided accordingly to each level, or equivalently the input coordinate is scaled up accordingly. Then by simply taking its floor and ceiling, we can identify its local parallelepiped and continue with our proposed simplex algorithm within the unit $n$-cube.

Additionally, due to coordinate skewing, the vertex $\textbf{1}_n^T$ is now $\sqrt{\textbf{n+1}}_n^T$ in the new coordinate system. The resulting parallelepiped is smaller than the original hypercube and cannot cover our sampling volume entirely. Therefore, we need to use an adjustment scale of $S_n = \sqrt{n+1}$ to avoid accessing points outside the simplicial grids.

\subsection{Hash Table Selection}

\begin{wrapfigure}{r}{0.32\textwidth}
\centering
\begin{tikzpicture}
    \draw (0, 0) node[inner sep=0] {\includegraphics[width=0.8\linewidth]{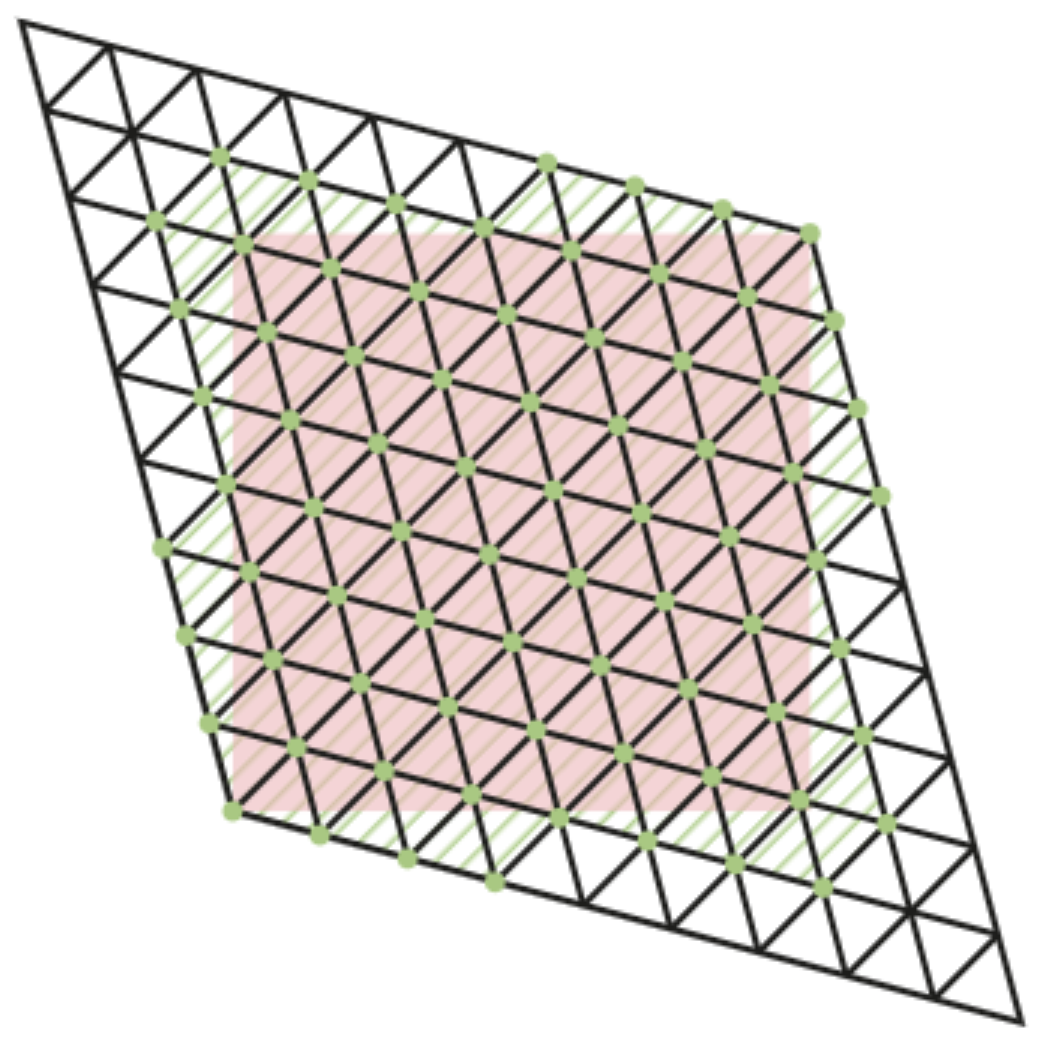}};
\end{tikzpicture}
\caption{{Only partial entries (colored in green) would be accessed and interpolated when sampling within the unit square (colored in red).}}
\label{fig:ratio}
\end{wrapfigure}

The choice of whether to use a dense grid or hash table can be task-specific for grid-based structures. However, using a dense grid to back a simplicial grid can be extremely inefficient. This is because only a portion of the vertices is accessed when we sample inside the unit $n$-cube, which leads to significant memory wastage if all vertex features are stored. Figure \ref{fig:ratio} provides a visual illustration of this in 2D. Assigning an order to the unused vertices to address this issue can cause unnecessary overhead, especially as it varies with grid size and dimension. To solve this problem, we have chosen to use a hash table. This approach avoids the need for explicit ordering, resulting in more efficient and scalable implementation of simplicial grids. 

To determine the size of the hash table when given dimension and level, we need to calculate the percentage of unused vertices. As the level increases, the volume covered by the used simplex vertices will converge to the volume of the hypercube. Assuming the grid is infinitely dense with at infinite level, the ratio of the two volumes would be 1. Then the percentage of vertices used in the simplex gird could be approximated by the volume ratio of the hypercube and the parallelpiped. The volume of the distorted $n$-cube, which is a parallelepiped as discussed in coordinate skewing, can be determined by $V = |\det(unskew(\boldsymbol{v_1}, \boldsymbol{v_2}, \ldots, \boldsymbol{v_n}))|$, where $\boldsymbol{v_1}, \boldsymbol{v_2}, \ldots, \boldsymbol{v_n}$ are the $n$-dimensional vectors that define the edges of the parallelepiped, and $|\cdot|$ denotes the absolute value of the determinant. In the skewed space, the vectors are on the axis and $(\boldsymbol{v_1}, \boldsymbol{v_2}, \ldots, \boldsymbol{v_n})$ is the identity matrix times $S_n$. By performing coordinate unskewing operation as Equation \ref{e3}, we can obtain the vector coordinates in the original space. Hence, $V=|S_n ^ n \det(I_n - G_n C_n)|$=$S_n ^ n (1-n G_n)$=$S_n^{n-1}$= $(n+1)^{\frac{n-1}{2}}$, where $I_n$ is the identity matrix and $C_n$ is the constant matrix of 1. Then, the ratio of the parallelepiped and the unit $n$-cube is $\frac{1}{V}$. As shown in Figure \ref{fig:ratio_dim}, this ratio decays exponentially. 

However, we would expect this ratio to be higher at lower levels due to discretization. We would like to derive an estimation that gives us a varying percentage at different levels. The result is reported in Figure \ref{fig:ratio_level}. As the level increases, the ratio quickly converges to the theoretical lower bound. Therefore, in practice, we could use a hash table with a size of the theoretical ratio to achieve similar quality compared to the collision-free implementation. We could also fix the hash table size and scale our level by $(n+1)^{\frac{n-1}{2n}}$, which is how we implemented it when compared with the baseline methods to guarantee equal memory size.

\begin{figure}[h]
\centering
\begin{subfigure}[b]{.48\textwidth}
    \resizebox{1.0\textwidth}{!}{%
        \begin{tikzpicture}
    	\begin{axis}[
            title={Theoretical memory usage},
            ymin = 0, ymax = 100,
    		xlabel=Dimension,
    		ylabel=Utilized Percentage,
            ylabel absolute, ylabel style={yshift=-0.3cm}]
        \addplot[mark=*, color=c_red] coordinates {
    		(2,57.7350)
    		(3,25.0000)
    		(4,8.9443)
    		(5,2.7778)
    		(6,0.7714)
    		(7,0.1953)
    	};
        \addplot [name path=upper,draw=none] table[x=x,y expr=\thisrow{max}] {data.dat};
        \addplot [name path=lower,draw=none] table[x=x,y expr=\thisrow{y}] {data.dat};
        \addplot [fill=c_red!30] fill between[of=upper and lower];
        \legend{L = $\infty$}
    	\end{axis}
        \end{tikzpicture}
    }%
    
\caption{{Theoretical percentage of the memory usage of simplex structures, which is $(n+1)^{-\frac{n-1}{2}}$. The volume ratio between the unit hypercube and the transformed parallelepiped is used for calculating the lower bound.}}
\label{fig:ratio_dim}
\end{subfigure}
\enskip
\begin{subfigure}[b]{.48\textwidth}
\resizebox{1.0\textwidth}{!}{%
\begin{tikzpicture}
	\begin{semilogxaxis}[
        xmin = 1, xmax = 256,
        ymin = 0, ymax = 100,
        title={Actual memory usage},
		xlabel=Level,
		ylabel=Utilized Percentage,
        ylabel absolute, ylabel style={yshift=-0.3cm}]
    \addplot[mark=*, color=c_red] coordinates {
		(2,100.0000)
		(4,92.0000)
		(8,75.3086)
		(16,68.8581)
		(32,63.4527)
		(64,60.7101)
		(128,59.2452)
	};
    \addplot[dashed, domain = 1:256, mark=none, color=c_red, samples=2, forget plot] {57.7350};
    \addplot[mark=*, color=c_green] coordinates {
		(2,55.5556)
		(4,52.0000)
		(8,40.7407)
		(16,33.6861)
		(32,29.5100)
		(64,27.3009)
		(128,26.1611)
	};
    \addplot[dashed, domain = 1:256, mark=none, color=c_green, samples=2, forget plot] {25.0000};
    \addplot[mark=*, color=c_blue] coordinates {
		(2,38.2716)
		(4,33.7600)
		(8,24.2189)
		(16,15.8870)
		(32,12.2095)
		(64,10.4383)
		(128,9.6328)
	};
    \addplot[dashed, domain = 1:256, mark=none, color=c_blue, samples=2, forget plot] {8.9443};
    \legend{n = 2, n = 3, n = 4}
	\end{semilogxaxis}
\end{tikzpicture}
}%
\caption{{Analytical percentage of the memory usage of simplex structures with different levels. The dashed lines correspond to the theoretical lower bounds in each dimension. Note that the results were calculated by sampling and may slightly underestimate.}}
\label{fig:ratio_level}
\end{subfigure}

\caption{{Analysis of memory usage for simplicial structures with different dimensions and levels.}}
\label{fig:ratio_all}
\end{figure}
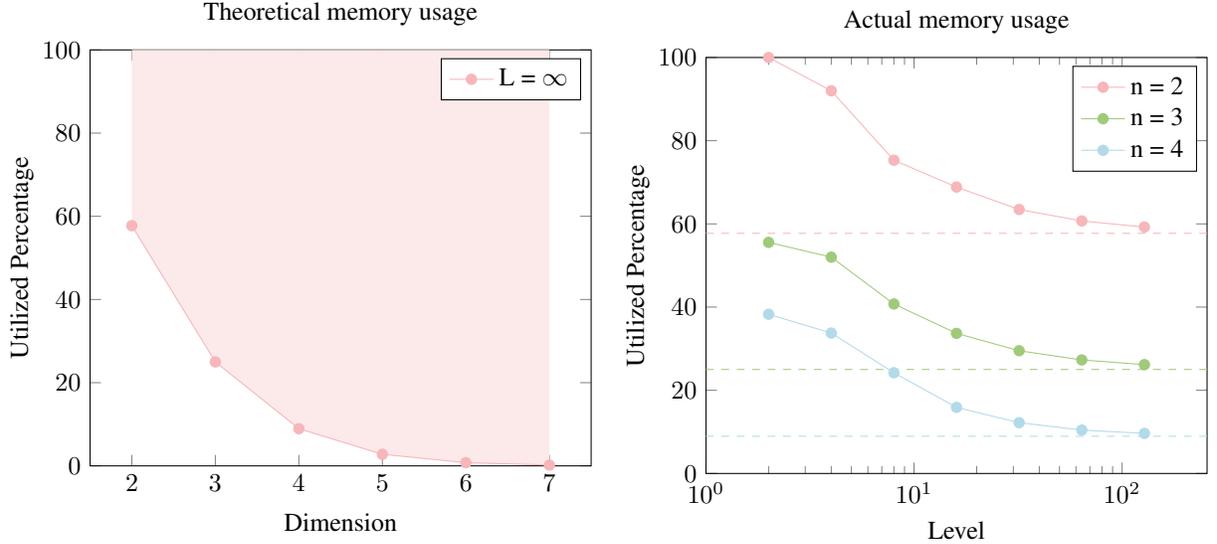

\section{Experiments}

To compare simplex-based and grid-based structures in multi-resolution hash encoding~\cite{mueller2022instant}, we employ a variety of tasks with increasing dimensional inputs to provide comprehensive results. Our evaluation process involves comparing the training performance and benchmark kernel run-time of dense-grid and simplex multi-resolution backbone methods. Through these experiments, we would showcase the versatility of the simplex-based structure and highlight its superiority over traditional grid-based encoding methods in various applications. We applied our method in the following tasks: 

\begin{enumerate}
\item \textbf{Gigapixel image}: the network learns the mapping from 2D coordinates to RGB colors of a high-resolution image.
\item \textbf{Volumetric Rendering}: the network learns the mapping from 5D coordinates to trace the rays in a given 3D space
\item \textbf{High dimensional analysis}: the network learns the mapping from $n$-dimension coordinates to the predetermined noise value.
\end{enumerate}

\subsection{Gigapixel Image}

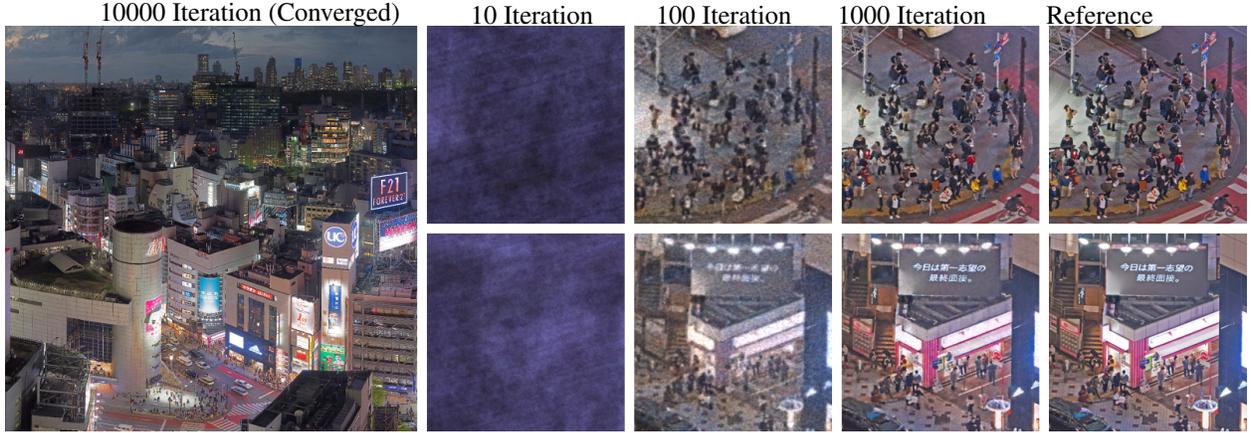
\begin{figure}[h]
\centering
\begin{tikzpicture}
    \draw (0, 0) node[inner sep=0] {\includegraphics[width=1.0\linewidth]{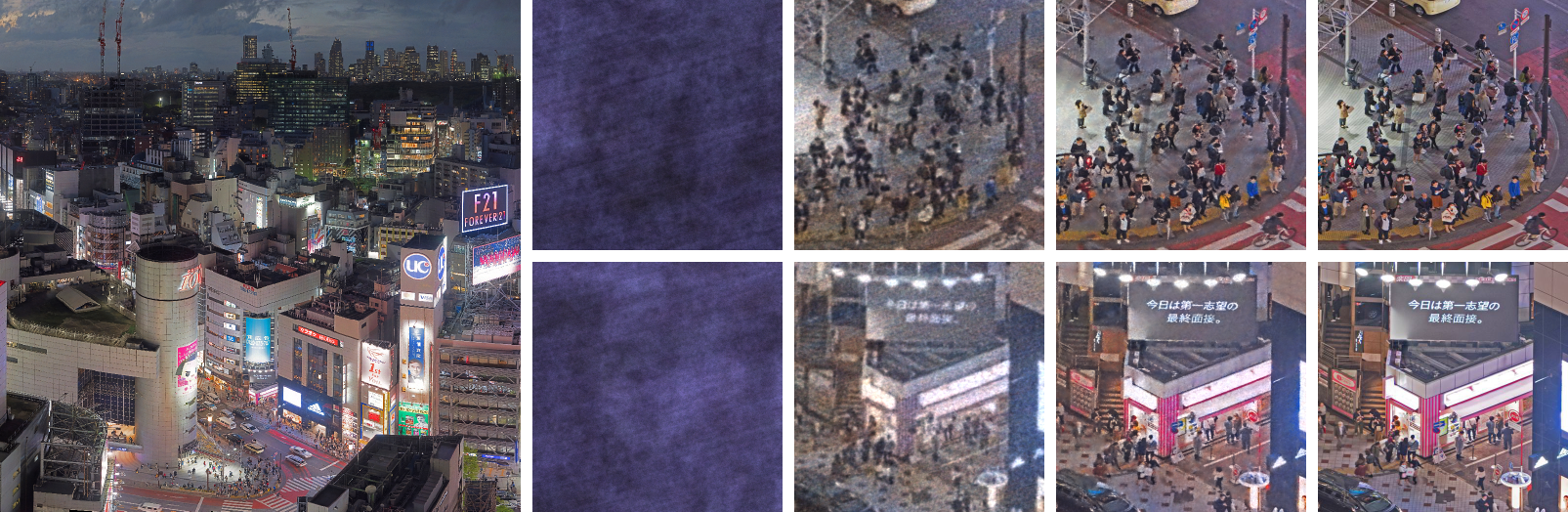}};
    \draw (-5.0, 2.85) node {10000 Iteration (Converged)};
    \draw (-1.25, 2.85) node {10 Iteration};
    \draw (1.3, 2.85) node {100 Iteration};
    \draw (3.8, 2.85) node {1000 Iteration};
    \draw (6.3, 2.85) node {Reference};
\end{tikzpicture}
\caption{{Optimization results from fitting an RGB image with 439M pixels (21450 $\times$ 21450). We use the same configurations with 7.9M trainable parameters (7.87M + 7k). Tokyo gigapixel photograph \textcopyright {Trevor Dobson \href{https://creativecommons.org/licenses/by-nc-nd/2.0/}{(CC BY-NC-ND 2.0)}}  }}
\label{fig:tokyo}
\end{figure}

Learning the 2D to RGB mapping of image coordinates to colors has become a popular benchmark for testing a model’s ability to represent high-frequency detail. Recent breakthroughs in adaptive coordinate networks have shown impressive results when fitting very large images—up to a billion pixels—with high fidelity at even the smallest scales~\cite{acorn, mueller2022instant}. We attempt to replicate the same experiment targeting to represent high-fidelity images with a hash table and implicit MLP in seconds to minutes. We begin by using the Tokyo gigapixel photograph as a reference image~\ref{fig:tokyo} and utilize simplex-based encoding to represent the image with hash maps and MLP parameters. Initially, the hash map features and the MLP weighting is randomly initialized so that the trained image appears noise-like. Through progressive weight back-propagation with ground truth and xy-to-RGB references appending, the network converges to the reference image with an indescribable difference. After 10,000 iterations, we are able to represent a 439M pixels image with only 7.9M trainable parameters, reaching a stunning 1.7 degrees of freedom in Figure~\ref{fig:tokyo}.

Practically, the simplex-based and grid-based encoding yielded close to almost identical runtime and PSNR scores for this image over-fitting task. With 16 multi-resolution levels and $2^19$ size of hash tables, after 10000 iterations, the simplex-based encoding obtains a PSNR of 29.94, whereas the grid-based encoding has a PSNR of 29.82. Additionally, 10000 iterations take 16.34 seconds for grid-based encoding and 14.94 seconds for simplex-based encoding. On 2D tasks like image overfitting, the grid-based structure requires information from 4 neighboring vertices, whereas the simplex-based structure has 3 neighboring vertices to interpolate with extra computational overhead on coordinate skewing and simplicial subdivision. While we do not expect any runtime improvement over instant-NGP, we would like to first use this experiment to verify the feasibility of simplex encoding. We use our analogous dense structure to check if we can produce similar results in both speed and quality. We then adopt the multi-resolution structure and compare the results with instant-NGP.

\subsection{Volumetric Rendering}
A more useful application is volumetric rendering, which computes the pixel color of a ray by integrating over transmittance and density. Given a ray vector $\textbf{r}$ parameterized by distance $t$ and viewing direction $\textbf{d}$, volumetric rendering computes its final pixel color $C(\textbf{r})$ by

\begin{equation} \label{e5}
    C(\textbf{r}) = \int_{t_n}^{t_f} T(t)\sigma(\textbf{r}(t))\textbf{c}(\textbf{r}(t),\textbf{d}) dt,\quad T(t) = \exp(-\int_{t_n}^t \sigma(\textbf{r}(s))ds),
\end{equation}

where $\sigma$ is the density and $T$ is the transmittance. Unlike raytracing and rasterizing, volumetric rendering is inherently differentiable which enabled us to learn 3D shape with only 2D supervision. This is first used for 3D reconstruction in NeRF~\cite{hedman2021snerg}, where the 3D scene is represented by a neural network. The network takes in 3D position and 2D direction and outputs the volume density of the particle at the position as well as RGB radiance at that position viewed from the given angle. To render an image, the network is queried multiple times at discrete points along the ray and their density and color are obtained. Using a volumetric rendering equation, the samples are composited into the final ray color. Finally, the L2 loss is computed based on ground truth pixel color and through gradient-descent, the network learns this 3D scene. 

\begin{figure}[ht]
\centering
\begin{subfigure}[b]{.48\textwidth}
    \resizebox{1.0\textwidth}{!}{%
        \begin{tikzpicture}
    	\begin{semilogxaxis}[
            title={Volumetric rendering kernel run-time},
            legend pos=north west,
            ymin = 0, ymax = 800,
    	xlabel=Number of Samples,
    	ylabel=Kernel run-time (s),
            ylabel absolute, ylabel style={yshift=-0.2cm}]
        \addplot[mark=*, color=c_red] coordinates {
    		(1,1.08509)
    		(4,2.27835)
    		(16,7.89389)
    		(64,28.75514)
    		(256,114.52051)
    		(1024,463.75629)
    	};
        \addplot[mark=*, color=c_green] coordinates {
                (1, 1.43159)
                (4, 3.42221)
                (16, 13.38685)
                (64, 50.04933)
                (256, 198.20725)
                (1024, 790.71214)
        };
        \addplot [name path=upper,draw=none] table[x=x,y expr=\thisrow{max}] {data.dat};
        \addplot [name path=lower,draw=none] table[x=x,y expr=\thisrow{y}] {data.dat};
        \legend{simplex, grid}
    	\end{semilogxaxis}
        \end{tikzpicture}
    }%
    
\caption{{The comparison between the volumetric rendering performance shows that simplex-based structure is consistently faster}}
\label{fig:ratio_dim}
\end{subfigure}
\enskip
\begin{subfigure}[b]{.48\textwidth}
\resizebox{1.0\textwidth}{!}{%
\begin{tikzpicture}
	\begin{semilogxaxis}[
        ymin = 1, ymax = 2,
        legend pos=north west,
        title={Kernel run-time ratio},
		xlabel=Number of samples,
		ylabel=grid/simplex kernel run-time ratio,
            ylabel absolute, 
            ylabel style={yshift=-0.3cm}]
    \addplot[mark=*, color=c_red] coordinates {
		(1,1.31932835064373)
		(4,1.50205631268242)
		(16,1.69584957479772)
		(64,1.74053508346682)
		(256,1.73075766078932)
		(1024,1.70501652926368)
	};
    \addplot [name path=upper,draw=none] table[x=x,y expr=\thisrow{max}] {data.dat};
    \addplot [name path=lower,draw=none] table[x=x,y expr=\thisrow{y}] {data.dat};
    \legend{ratio}
	\end{semilogxaxis}
\end{tikzpicture}
}%
\caption{{The grid-to-simplex run-time ratio plateaus at 1.8 - approaches the theoretical upper bound at 3D}}
\label{fig:ratio_level}
\end{subfigure}

\caption{{Analysis of memory usage for simplicial structures with different dimensions and levels.}}
\label{fig:volume_all}
\end{figure}
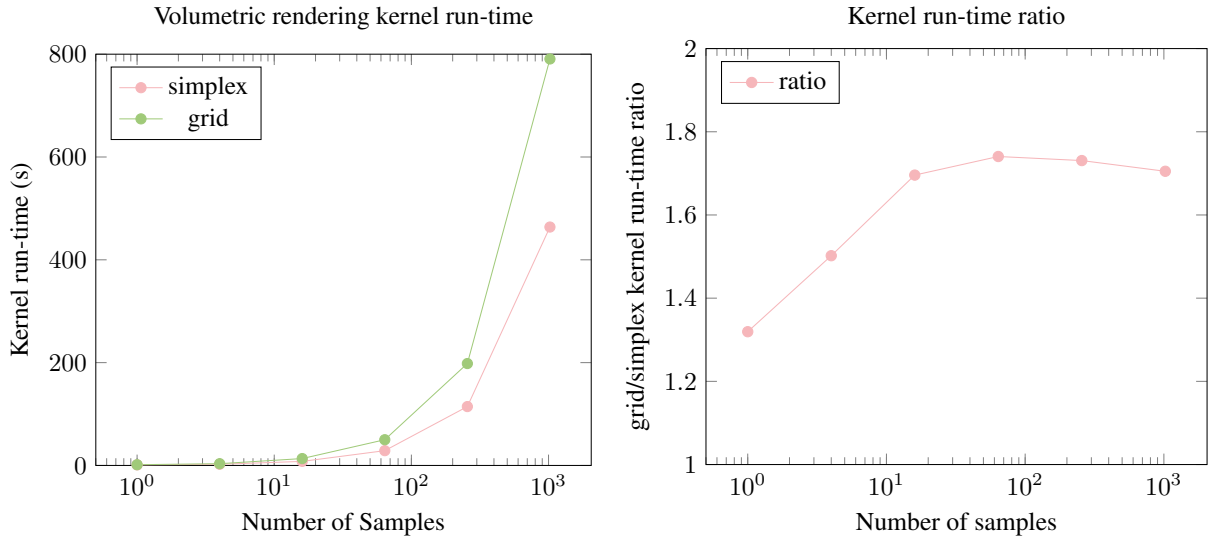

Note that in the volumetric rendering equation, transmittance T in Equation~\ref{e5} depends on previously sampled densities and cannot be evaluated in parallel. This makes our implementation extremely efficient because as the number of samples increases, the runtime of the entire algorithm would approach the theoretical bounds, which is a 2-time speed up in 3D (with 4 vertices in simplex vs 8 vertices in the grid). Through repetitive sampling, our computational overhead for sorting coordinates for each vertex in the Simplicial Subdivision phase (Refer to \textbf{Section 4.2}) becomes negligible. This effect is proven in Figure~\ref{fig:volume_all}. This can allow faster training and rendering for NeRF without any loss in quality.

\subsection{Kernel Analysis}

To investigate the performance of our core implementation, we compare the kernel run time with baseline implementation on both CPU and GPU. For the CPU, we used an Intel Core i7-8700K CPU with 6 cores and 12 threads, running at 2.6 GHz with 16 GB of RAM (2019 Macbook Pro 16 inch). We implemented both the baseline method (grid implementation) and our proposed method (simplex hash implementation) on this CPU and measured the kernel run-time of both methods for various dimensional inputs. We used c++ for our implementation and measured the kernel run-time using the chrono module. 

For each dimension, we use $2^{27}$ cells and randomly sample $2^{10}$ data points inside the n-dimensional structure. In order to produce a result in the seconds level, we perform the computation for each method 1000 times. Note that the side length of the grid is $\sqrt[n]{2^{27}}$. For a 3-dimensional input, for example, the side length is $\sqrt[3]{2^{27}}$ = $2^9$. For each dimension, we run the experiments 5 times to calculate the average of the kernel run-time for each method. The experiment result is summarized in Figure \ref{fig:ratio_all}.

\begin{figure}[ht]
\centering
\begin{subfigure}[b]{.48\textwidth}
    \resizebox{1.0\textwidth}{!}{%
        \begin{tikzpicture}
    	\begin{axis}[
            legend pos=north west,
            title={CPU Kernel Run-time},
            ymin = 0, ymax = 80,
    		xlabel=Dimension,
    		ylabel=Kernel run-time (s),
            ylabel absolute, ylabel style={yshift=-0.2cm}]
        \addplot[mark=*, color=c_red] coordinates {
    		(2,3.91020)
    		(3,4.94373)
    		(4,5.94416)
    		(5,7.09993)
    		(6,8.26581)
    		(7,8.61585)
    	};
        \addplot[mark=*, color=c_green] coordinates {
                (2, 4.10982)
                (3, 6.29560)
                (4, 10.06391)
                (5, 16.18120)
                (6, 32.30852)
                (7, 69.09387)
        };
        \addplot [name path=upper,draw=none] table[x=x,y expr=\thisrow{max}] {data.dat};
        \addplot [name path=lower,draw=none] table[x=x,y expr=\thisrow{y}] {data.dat};
        \legend{simplex, grid} 
    	\end{axis}
        \end{tikzpicture}
    }%
    
\caption{{Simplex run-time scales linearly with dimension, while grid run-time scales exponentially}}
\label{fig:ratio_dim}
\end{subfigure}
\enskip
\begin{subfigure}[b]{.48\textwidth}
\resizebox{1.0\textwidth}{!}{%
\begin{tikzpicture}
	\begin{axis}[
        ymin = 0, ymax = 150,
        legend pos=north west,
        title={Vertex Count},
		xlabel=Dimension,
		ylabel=Number of vertices,
            ylabel absolute, 
            ylabel style={yshift=-0.2cm}]
    \addplot[mark=*, color=c_red] coordinates {
		(2,3)
		(3,4)
		(4,5)
		(5,6)
		(6,7)
		(7,8)
	};
    \addplot[mark=*, color=c_green] coordinates {
		(2,4)
		(3,8)
		(4,16)
		(5,32)
		(6,64)
		(7,128)
	};
    \addplot [name path=upper,draw=none] table[x=x,y expr=\thisrow{max}] {data.dat};
    \addplot [name path=lower,draw=none] table[x=x,y expr=\thisrow{y}] {data.dat};
    \legend{simplex, grid}
	\end{axis}
\end{tikzpicture}
}%
\caption{{Simplex vertices scales linearly with dimension, while grid vertices scales exponentially}}
\label{fig:ratio_level}
\end{subfigure}

\caption{{Analysis of memory usage for simplicial structures with different dimensions and different levels. Both graph exhibits the same pattern with dimension}}
\label{fig:ratio_all}
\end{figure}
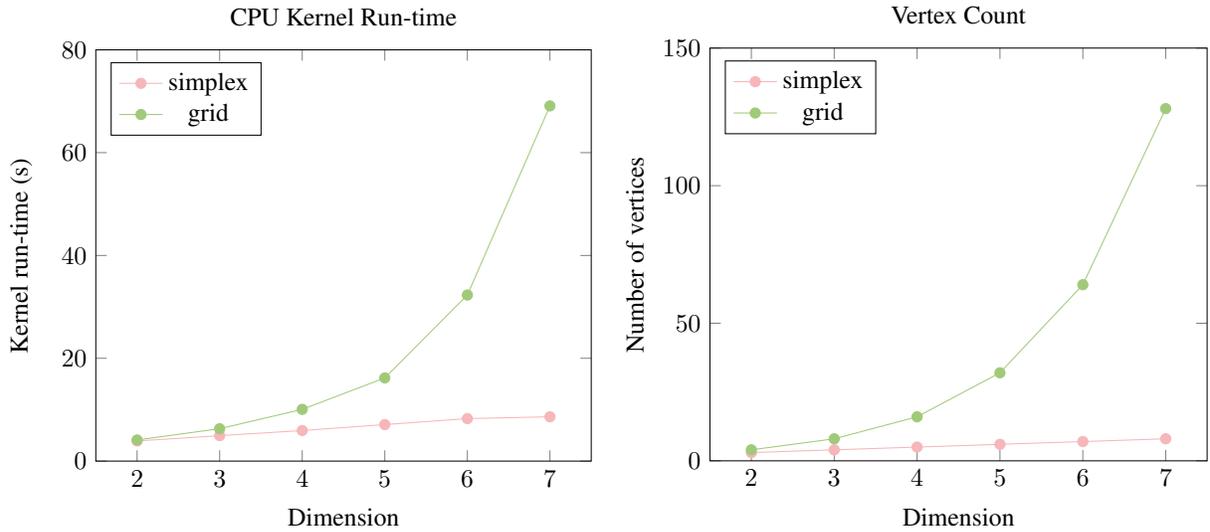

According to the graph, the kernel run-time of simplices scales much better with dimension. The simplex-based encoding scales linearly because its number of vertices also scales linearly with dimension. On the other hand, the kernel run-time for grid-based structure scales exponentially - matches our observation on the exponential growth of the number of vertices with respect to dimension. This gives us a huge competitive advantage against grid-based encoding in high-dimensional tasks such as NeRF and SDF.

\section{Discussion}
In this paper, we present a new approach for parameterizing graphics primitives using a simplex-based structure that offers significant advantages over traditional algorithms. By representing primitives as simplices, we are able to reduce the memory access and interpolation complexity, resulting in more efficient implementation. Through repetitive experiments and benchmarking, we show that our approach scales exceptionally well with dimensionality, making it particularly well-suited for tasks such as volumetric rendering. We believe that the simplicity, efficiency, and versatility of our approach make it an exciting avenue for future research in graphics primitives and beyond.

\bibliographystyle{unsrt}
\newpage \bibliography{references}

\begin{thebibliography}{10}

\bibitem{Occupancy_Networks}
Lars Mescheder, Michael Oechsle, Michael Niemeyer, Sebastian Nowozin, and Andreas Geiger.
\newblock Occupancy networks: Learning 3d reconstruction in function space.
\newblock In {\em Proceedings IEEE Conf. on Computer Vision and Pattern Recognition (CVPR)}, 2019.

\bibitem{chen2018implicit_decoder}
Zhiqin Chen and Hao Zhang.
\newblock Learning implicit fields for generative shape modeling.
\newblock {\em Proceedings of IEEE Conference on Computer Vision and Pattern Recognition (CVPR)}, 2019.

\bibitem{Park_2019_CVPR}
Jeong~Joon Park, Peter Florence, Julian Straub, Richard Newcombe, and Steven Lovegrove.
\newblock Deepsdf: Learning continuous signed distance functions for shape representation.
\newblock In {\em The IEEE Conference on Computer Vision and Pattern Recognition (CVPR)}, June 2019.

\bibitem{Vicini2022sdf}
Delio Vicini, Sébastien Speierer, and Wenzel Jakob.
\newblock Differentiable signed distance function rendering.
\newblock {\em Transactions on Graphics (Proceedings of SIGGRAPH)}, 41(4):125:1--125:18, July 2022.

\bibitem{mildenhall2020nerf}
Ben Mildenhall, Pratul~P. Srinivasan, Matthew Tancik, Jonathan~T. Barron, Ravi Ramamoorthi, and Ren Ng.
\newblock Nerf: Representing scenes as neural radiance fields for view synthesis.
\newblock In {\em ECCV}, 2020.

\bibitem{SunSC22}
Cheng Sun, Min Sun, and Hwann{-}Tzong Chen.
\newblock Direct voxel grid optimization: Super-fast convergence for radiance fields reconstruction.
\newblock In {\em CVPR}, 2022.

\bibitem{Lombardi:2019}
Stephen Lombardi, Tomas Simon, Jason Saragih, Gabriel Schwartz, Andreas Lehrmann, and Yaser Sheikh.
\newblock Neural volumes: Learning dynamic renderable volumes from images.
\newblock {\em ACM Trans. Graph.}, 38(4):65:1--65:14, July 2019.

\bibitem{hedman2021snerg}
Peter Hedman, Pratul~P. Srinivasan, Ben Mildenhall, Jonathan~T. Barron, and Paul Debevec.
\newblock Baking neural radiance fields for real-time view synthesis.
\newblock {\em ICCV}, 2021.

\bibitem{yu_and_fridovichkeil2021plenoxels}
{Sara Fridovich-Keil and Alex Yu}, Matthew Tancik, Qinhong Chen, Benjamin Recht, and Angjoo Kanazawa.
\newblock Plenoxels: Radiance fields without neural networks.
\newblock In {\em CVPR}, 2022.

\bibitem{yu2021plenoctrees}
Alex Yu, Ruilong Li, Matthew Tancik, Hao Li, Ren Ng, and Angjoo Kanazawa.
\newblock {PlenOctrees} for real-time rendering of neural radiance fields.
\newblock In {\em ICCV}, 2021.

\bibitem{Chen2022ECCV}
Anpei Chen, Zexiang Xu, Andreas Geiger, Jingyi Yu, and Hao Su.
\newblock Tensorf: Tensorial radiance fields.
\newblock In {\em European Conference on Computer Vision (ECCV)}, 2022.

\bibitem{cao2023hexplane}
Ang Cao and Justin Johnson.
\newblock Hexplane: a fast representation for dynamic scenes.
\newblock {\em arXiv preprint arXiv:2301.09632}, 2023.

\bibitem{Chan2021}
Eric~R. Chan, Connor~Z. Lin, Matthew~A. Chan, Koki Nagano, Boxiao Pan, Shalini~De Mello, Orazio Gallo, Leonidas Guibas, Jonathan Tremblay, Sameh Khamis, Tero Karras, and Gordon Wetzstein.
\newblock Efficient geometry-aware {3D} generative adversarial networks.
\newblock In {\em arXiv}, 2021.

\bibitem{fridovich2023k}
Sara Fridovich-Keil, Giacomo Meanti, Frederik Warburg, Benjamin Recht, and Angjoo Kanazawa.
\newblock K-planes: Explicit radiance fields in space, time, and appearance.
\newblock {\em arXiv preprint arXiv:2301.10241}, 2023.

\bibitem{mueller2022instant}
Thomas M\"uller, Alex Evans, Christoph Schied, and Alexander Keller.
\newblock Instant neural graphics primitives with a multiresolution hash encoding.
\newblock {\em ACM Trans. Graph.}, 41(4):102:1--102:15, July 2022.

\bibitem{perlin_noise}
Ken Perlin.
\newblock An image synthesizer.
\newblock {\em SIGGRAPH Comput. Graph.}, 19(3):287–296, jul 1985.

\bibitem{simplex_noise}
Ken Perlin.
\newblock Chapter 2 noise hardware.

\bibitem{acorn}
Julien N.~P. Martel, David~B. Lindell, Connor~Z. Lin, Eric~R. Chan, Marco Monteiro, and Gordon Wetzstein.
\newblock Acorn: Adaptive coordinate networks for neural scene representation.
\newblock 2021.

\end{thebibliography}

\newpage \section{Appendix}

\subsection*{Appendix A. Mathematical proofs to simplex-based structures}
\begin{theorem}\label{t1}
$S=\{ \textbf{x}\in\mathbb{R}^n : 0\leq x_1\leq \cdots \leq x_n\leq 1\}$ is a $n$-simplex.
\end{theorem}

\begin{proof}
Let $M$ be an upper triangular $(n+1)\times (n+1)$ matrix with only 0 and 1. Then for any $\textbf{x}\in S$, the solution to

\begin{equation} \label{e6} 
\begin{bmatrix} 
    1 & \cdots  & 1 & 1\\
    0 & 1 & \cdots  & 1\\
    \vdots & \ddots & \ddots & \vdots\\
    0 & \cdots & 0  & 1 
\end{bmatrix} \boldsymbol{\lambda} = 
        \begin{bmatrix}
           1 \\
           x_{n} \\
           \vdots \\
           x_{1}
         \end{bmatrix},\quad \boldsymbol{\lambda} \in \mathbb{R}^{n+1},
\end{equation}

should be $\boldsymbol{\lambda}= [1-x_n,x_n-x_{n-1},\cdots,x_2-x_1,x_1]^T$. Consider points $\textbf{v}_1, \cdots, \textbf{v}_{n+1}$, where the first $i+1$ entries of $\textbf{v}_i$ are 1 and the rest are 0. Then $M$ can also be expressed as 
$\begin{bmatrix} 
    1 & \cdots & 1 \\
    \textbf{v}_1 & \cdots & \textbf{v}_{n+1}
\end{bmatrix}$
. The solution shows that $\textbf{x}$ can be written as a linear combination of the $n+1$ points. Addi tionally, given that $0\leq x_1\leq \cdots \leq x_n\leq 1$, every entry of $\boldsymbol{\lambda}$ is no less than 0. With the addition constraint $\sum_i{\lambda_i}=1$, we conclude that $\textbf{x}$ can be written as a convex combination of the $n+1$ points and hence is inside the convex hull $C$ of points $\textbf{v}_1, \cdots, \textbf{v}_{n+1}$. Therefore, we have $C \subseteq S$. 

Let $\textbf{x}^{\prime} \in C$. Therefore, there exists such $\boldsymbol{\lambda} \in \mathbb{R}^{n+1}$ s.t. 

\begin{equation} \label{e6} 
\begin{bmatrix} 
    \vert & \cdots  & \vert\\
    \textbf{v}_1 & \cdots & \textbf{v}_{n+1}\\
    \vert & \cdots & \vert 
\end{bmatrix} \boldsymbol{\lambda} = \textbf{x}^{\prime},\quad \sum_{i=1}^{n+1} {\lambda_i} = 1,\quad \lambda_i\geq 0 \text{ for all } i.
\end{equation}

Then, we have $x_j^{\prime} = \sum_{j+1}^{n+1} {\lambda_j}$, which indicates that $0\leq x^{\prime}_1\leq \cdots \leq x^{\prime}_n\leq 1$. Therefore, we have $\textbf{x}^{\prime} \in S$ and hence $S \subseteq C$. 

In conclusion, $S=C$. Since the $n+1$ vertices $\textbf{v}_1, \cdots, \textbf{v}_{n+1}$ are affinely independent points, its convex hull $C$ is a $n$-simplex and so is $S$.
\end{proof}

\begin{remark}\label{t2}
Let $\pi$ denote a permutation of $\{1,\cdots,n\}$, then $S_{\pi}=\{ \textbf{x}\in\mathbb{R}^n : 0\leq x_{\pi(1)}\leq \cdots \leq x_{\pi(n)}\leq 1\}$ is a $n$-simplex. The $n$-simplex has $n+1$ vertices $\textbf{v}_1, \cdots, \textbf{v}_{n+1}$, where all entries at indices $\{\pi_1,\cdots,\pi_{i-1}\}$ of $\textbf{v}_i$ are 1 and the rest are 0. Additionally, $\textbf{v}_1$ is $\textbf{0}^T$ and $\textbf{v}_{n+1}$ is $\textbf{1}^T$.
\end{remark}

\begin{theorem}\label{t3}
All possible $S_{\pi}$ are congruent. 
\end{theorem}

\begin{proof}
For any $S_{\pi}$, consider its $n+1$ vertices $\textbf{v}_1, \cdots, \textbf{v}_{n+1}$ as defined in Remark~\ref{t2}. Any two vertices of $S_{\pi}$ is an edge and it has $\frac{n(n+1)}{2}$ edges with different lengths. Let $d_k$ denote the distance between two difference vertices $\textbf{v}_i$ and $\textbf{v}_j$, where $i>j$ and $k=i-j$. As shown in Remark~\ref{t2}, consecutive vertices only differ by 1 at one entry, and we have $d_1=\sqrt{1}=1$. Similarly, vertices that differ by $k$ in their order have $k$ more (or less) 1 and $d_k=\sqrt{k}$. Therefore, for any $S_{\pi}$, it contains $n+1-k$ edges with length $\sqrt{k}$ where $k$ is from $1$ to $n$. Since all simplices have the same edges with each other, they are congruent regardless of the order of $\pi$. 
\end{proof}

\begin{lemma}\label{l1}
All possible $S_{\pi}^{\prime}$ in the transformed coordinate system are still congruent in the original coordinate system. 
\end{lemma}

\begin{proof}
We first obtain the vertex coordinates of $S_{\pi}^{\prime}$ in the original coordinate system using coordinate unskewing in Equation~\ref{e3}. For its vertex $\textbf{v}_i^{\prime}$, every entry is subtracted by $(i-1)G_n$. Using similar notation as above, $d_1^{\prime}=\sqrt{(1-G_n)^2 + (n-1)G_n^2}$. Similarly, $d_k^{\prime}=\sqrt{k(1-kG_n)^2 + (n-k)(kG_n)^2}$. Therefore, for any $S_{\pi}^{\prime}$, it contains $n+1-k$ edges with length $d_k$ where $k$ is from $1$ to $n$. Since all transformed simplices still have the same edges with each other, they are congruent regardless of the order of $\pi$.
\end{proof}

\begin{theorem}\label{t4}
A $n$-cube can be triangulated into $n!$ disjoint congruent simplices.
\end{theorem}

\begin{proof}
Based on Remark~\ref{t2}, the hypercube $[\textbf{0}, \textbf{1}]^n$ fully contains all $S_{\pi}$ where $\pi$ ranges over all possible $n!$ permutations of $\{1,2,...,n\}$.

Assume, to the contrary, that two different simplices $S_{\pi}, S_{\pi^*}$ with their two corresponding permutations $\pi, \pi^*$ intersect each other. Then $\exists \textbf{x}\in \mathbb{R}^n,$ s.t. $\textbf{x}$ is strictly in the interior of both $S_{\pi}$ and $S_{\pi^*}$. By sorting the entries of $\textbf{x}$, if the order satisfy both constraints from $S_{\pi}, S_{\pi^*}$, there must exist two entries with the same value. Therefore as the inequality constraints are not strictly satisfied, $\textbf{x}$ has to be on the surface of both $S_{\pi}, S_{\pi^*}$. Contradiction.

Since there are $n!$ such permutations, there are $n!$ simplices with the disjoint interior contained by the hypercube. Together with Lemma~\ref{l1}, the hypercube can be triangulated into $n!$ disjoint congruent simplices.
\end{proof}

\subsection*{Appendix B. Trilinear and barycentric interpolation}

\begin{algorithm}
\caption{Trilinear Inpoterlation}
\begin{algorithmic}[1]
\Function{Trilinear Interpolation}{$x$, $points$}
    \State $n \gets length(points)$
    \If{$n = 1$}
        \State \Return $points[0].value$
    \EndIf
    \State $i \gets 0$
    \While{$i < n - 1$ and $x > points[i + 1]$}
        \State $i \gets i + 1$
    \EndWhile
    \State $t \gets (x - points[i]) / (points[i + 1] - points[i])$
    \State $y0 = Trilinear Interpolation(x, points[0:i+2])$
    \State $y1 = Trilinear Interpolation(x, points[i:n])$  
    \Comment{Continue calculation in n-1 dimension}
    \State $\Return$ $y0 * (1 - t) + y1 * t$
\EndFunction
\end{algorithmic}
\end{algorithm}

\begin{algorithm}
\caption{Barycentric Interpolation}\label{alg:kernel}
\begin{algorithmic}
\For i in n+1 dimensions
\State $F_n = \frac{\sqrt{n+1}-1}{n}$
\State $x_i = x_i + 1_i^T \cdot F \sum_i{x_i}$ \Comment{Coordinate Skewing}
\State start-weight = 1
\State initialize w = $w_1, w_2, w_3, ... w_n$
\State f = bubble-sort($f_1, f_2, f_3, ... f_n$)
\State $\lfloor x \rfloor$ = Floored coordinates $x_1, x_2, x_3, ... x_n$
\If{i = 0}
\State $coordinate_i$ = the index of largest coordinate in $\lfloor x \rfloor$
\State $w_i$ = start-weight - $coordinate_i$
\State start-weight = $f_1$
\Else
\State $coordinate_i$ = the index of i-th largest coordinate in $\lfloor x \rfloor$
\State $x_{coordinate_i} $ += 1
\State $w_i$ = start-weight - $coordinate_i$ \Comment{Simplicial Subdivision}
\EndIf
\EndFor
\State g = 2 features for point \textbf{x}
\State x-feature = $w \cdot g$ \Comment{Return feature for point x}
\end{algorithmic}
\end{algorithm}






\end{document}